\documentclass[conference]{IEEEtran}
\IEEEoverridecommandlockouts
\usepackage{cite}
\usepackage{amsmath,amssymb,amsfonts}
\usepackage{algorithmic}
\usepackage{graphicx}
\usepackage{textcomp}
\usepackage{xcolor}

\usepackage{amsmath}
\usepackage{multirow}
\usepackage{graphicx}
\usepackage[normalem]{ulem}
\useunder{\uline}{\ul}{}
\usepackage{array}
\usepackage{float}
\usepackage{caption}

\def\BibTeX{{\rm B\kern-.05em{\sc i\kern-.025em b}\kern-.08em
    T\kern-.1667em\lower.7ex\hbox{E}\kern-.125emX}}
\begin{document}

\title{Unimodal-driven Distillation in Multimodal Emotion Recognition with Dynamic Fusion \thanks{\textsuperscript{*}~Huaicheng Yan is the Corresponding author (Email: hcyan@ecust.edu.cn).}}

% \author{Anonymous ICME submission}

\author{
\IEEEauthorblockN{1\textsuperscript{st} Jiagen Li}
\IEEEauthorblockA{
    \textit{dept. School of Information Science and Engineering} \\
    \textit{East China University of Science and Technology}\\
    Shanghai, China\\
    y80220334@mail.ecust.edu.cn
}
\vspace{1em}
\IEEEauthorblockN{3\textsuperscript{rd} Huihao Huang}
\IEEEauthorblockA{
    \textit{dept. School of Mathematics and Informatics} \\
    \textit{South China Agricultural University} \\
    Guangzhou, China\\
    3h@stu.scau.edu.cn
}
\and
\IEEEauthorblockN{2\textsuperscript{nd} Rui Yu}
\IEEEauthorblockA{
    \textit{dept. School of Information Science and Engineering} \\
    \textit{East China University of Science and Technology}\\
    Shanghai, China\\
    y80220166@mail.ecust.edu.cn
}
\vspace{1em}
\IEEEauthorblockN{4\textsuperscript{th} Huaicheng Yan\textsuperscript{*}}
\IEEEauthorblockA{
    \textit{dept. School of Information Science and Engineering} \\
    \textit{East China University of Science and Technology} \\
    Shanghai, China\\
    hcyan@ecust.edu.cn\textsuperscript{*}
}
}

\maketitle

\begin{abstract}
    Multimodal Emotion Recognition in Conversations (MERC) identifies emotional states across text, audio and video, which is essential for intelligent dialogue systems and opinion analysis. Existing methods emphasize heterogeneous modal fusion directly for cross-modal integration, but often suffer from disorientation in multimodal learning due to modal heterogeneity and lack of instructive guidance. In this work, we propose SUMMER, a novel heterogeneous multimodal integration framework leveraging Mixture of Experts with Hierarchical Cross-modal Fusion and Interactive Knowledge Distillation. Key components include a Sparse Dynamic Mixture of Experts (SDMoE) for capturing dynamic token-wise interactions, a Hierarchical Cross-Modal Fusion (HCMF) for effective fusion of heterogeneous modalities, and Interactive Knowledge Distillation (IKD), which uses a pre-trained unimodal teacher to guide multimodal fusion in latent and logit spaces. Experiments on IEMOCAP and MELD show SUMMER outperforms state-of-the-art methods, particularly in recognizing minority and semantically similar emotions.
\end{abstract}

\begin{IEEEkeywords}
    Multimodal, Emotion Recognition, Knowledge Distillation, Mixture of Experts (© 2025 lEEE.)
\end{IEEEkeywords}

\section{Introduction}
\begin{figure}[t]
        \centering
        \includegraphics[width=\linewidth]{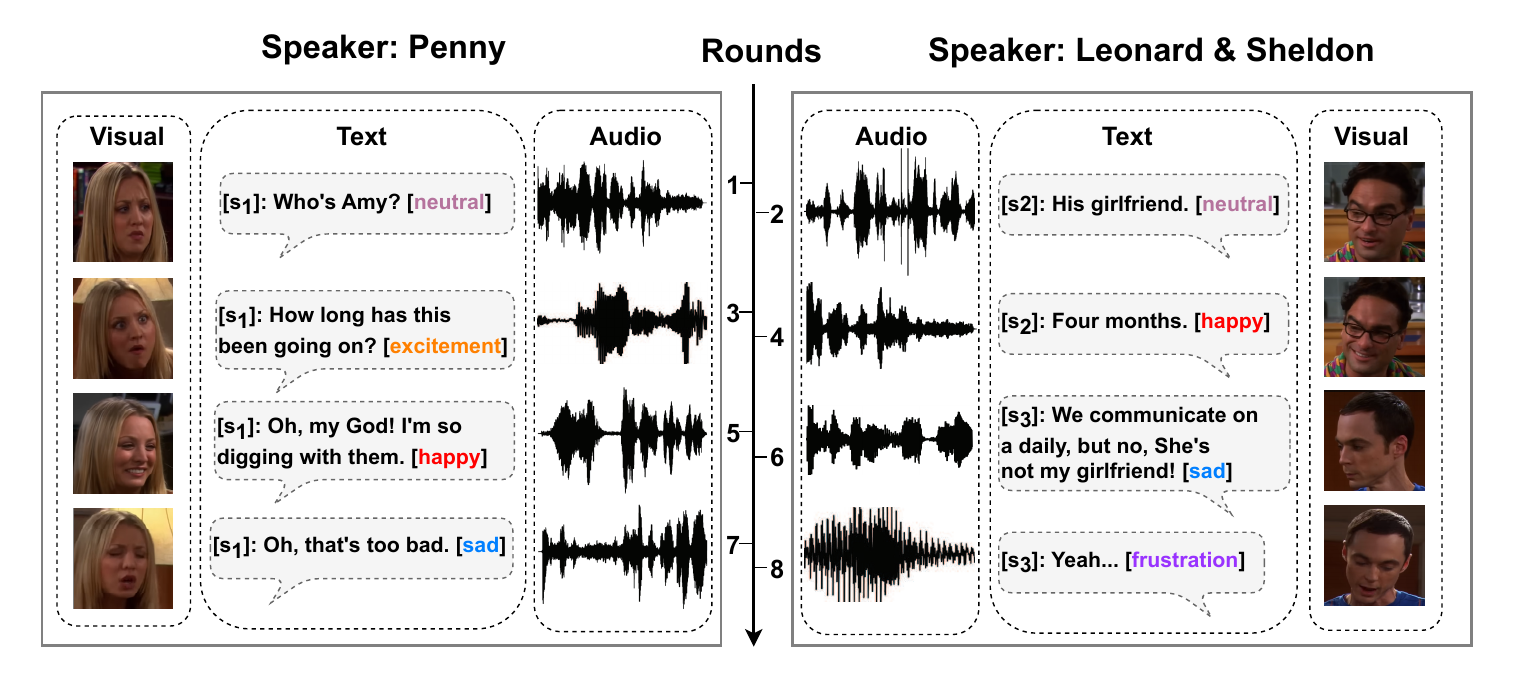}
        \caption{A representative example of multimodal emotion recognition in conversations from The Big Bang Theory.}
        \label{fig:Intro}
    \end{figure}

% \label{sec:intro}
    Multimodal Emotion Recognition in Conversations (MERC) enhances human-computer interaction and empathy in domains like digital humans, healthcare, and social media analytics. By integrating text, audio and visual cues (Fig. \ref{fig:Intro}), MERC captures nuanced emotional dynamics and enables adaptive feedback across diverse contexts.

    In MERC tasks, existing studies primarily focus on global context modeling and cross-modal fusion. Transformer-based models like MultiEMO and SDT \cite{MultiEMO, SDT} utilize attention mechanisms and knowledge distillation to capture long-range dependencies and integrate multimodal contextual information. However, challenges in modal association remain, notably in two aspects: (a) Traditional attention mechanisms exhibit limitations in global context modeling, whereas the Mixture of Experts (MoE) model enhances performance by leveraging a gating network, enabling resource allocation to critical features and improving contextual understanding. By activating only a few experts, MoE significantly reduces computational overhead compared to fully connected networks, particularly in large-scale multimodal tasks. However, its dependency on a fixed Top-K subset restricts adaptability in complex MERC scenarios, underscoring the necessity for dynamic token-wise selection to optimize contextual modeling and intra-modal integration. (b) Existing models often directly fuse multimodal features without addressing the inherent heterogeneity between modalities, leading to misalignment in feature representations, while cross-modality distillation facilitates knowledge transfer across heterogeneous features. However, it often suffers from multimodal learning disorientation and representation gaps in sentiment analysis. The guidance of a well-pretrained teacher model is essential for bridging these gaps and enhancing the efficacy of multimodal integration.
    
    To address these issues, we propose \textbf{SUMMER} (\textbf{S}parse \textbf{U}nimodal-driven distillation for \textbf{M}ulti-\textbf{M}odal \textbf{E}motion \textbf{R}ecognition), a framework to address challenges in intra- and inter-modal fusion and disorientation learning. First, we introduce a Sparse Dynamic Mixture of Experts (SDMoE) which excels in identifying and focusing on the most relevant features, improving contextual understanding and intra-modal integration. Moreover, dynamic token-wise selection in SDMoE enables adaptability to varying levels of complexity within different MERC scenarios, ensuring more precise and robust performance. Next, we present Hierarchical Cross-Modal Fusion (HCMF) to capture and integrate intrinsic inter-modal relationships, thereby improving global context understanding within inter-modal. Furthermore, we propose Interactive Knowledge Distillation (IKD), where a high-performing unimodal teacher model enhances the multimodal student model by aligning heterogeneous modalities, reducing feature distribution gaps, and providing rich supervision through soft labels that capture inter-class relationships.
    
    The primary contributions of this work are as follows:
    \begin{itemize}
      \item A novel Sparse Dynamic Mixture of Experts is proposed to enhance token-wise intra-modal selection, along with a Hierarchical Cross-Modal Fusion to refine heterogeneous inter-modal clues. This approach improves both local and global modeling for more effective fusion.\item A retrograde Knowledge Distillation strategy is introduced, leveraging a lightweight unimodal-driven teacher model to guide the multimodal student model, effectively addressing disorientation in multimodal learning.\item Our proposed framework achieves state-of-the-art performance on the IEMOCAP and MELD datasets, demonstrating exceptional capability in capturing subtle emotional nuances and excelling in recognizing semantically similar and underrepresented emotion categories.
    \end{itemize}

\section{Related Work}
\subsection{Multimodal Emotion Recognition in Conversations}
        MERC aims to analyze speakers' emotions by integrating intra- and inter-modal interactions from multimodal data. Advanced methods utilize attention mechanisms to enhance cross-modal encoding and sentiment trend. For instance, CTNet \cite{CTNet} employs cross-modality Transformers, while CKETF \cite{CKETF} improves context and knowledge representation. Emocaps \cite{emocaps} refine attention based emotion capsule to extract features. Tailor \cite{tailor} integrates unimodal extraction with multi-label decoding to capture label and modality dependencies while SDT \cite{SDT} introduces self-distillation for emotional interactions. 
        % Our model incorporates dynamic MoE and cross-modal integration to achieve effective intra- and inter-modal fusion.

    \subsection{Knowledge Distillation}
        Knowledge Distillation (KD) enhances efficiency by transferring knowledge from a larger teacher model to a smaller student model. In MERC tasks, KD enables student models to capture richer emotional representations for cross-modal integration. SENet \cite{SENet} transfers visual knowledge to speech, while Schoneveld \cite{Schoneveld} applies KD to facial expression recognition. FASD \cite{fasd} achieves heterogeneous model distillation by adaptively unifying voxel features. However, most methods rely on offline distillation with multimodal teachers, overlooking unimodal-driven approaches for complex multimodal students. To address this gap, our work focuses on efficient cross-modal learning through unimodal-driven knowledge transfer.

\section{Methodology}
\subsection{Task Definition}
    In MERC tasks, a conversation comprises $n$ utterances $\{u_1, u_2, ..., u_n\}$ and $m$ speakers $\{s_1, s_2, ..., s_m\}$. Each utterance $u_i$ encompasses three modalities: text($u_i^t$), audio($u_i^a$), and visual($u_i^v$). The goal is to predict the sentiment label $y_i$ for each utterance $u_i$ within the conversation.

\subsection{Model Overview}
\begin{figure*}[t]
    \centering
    \includegraphics[width=0.9\textwidth]{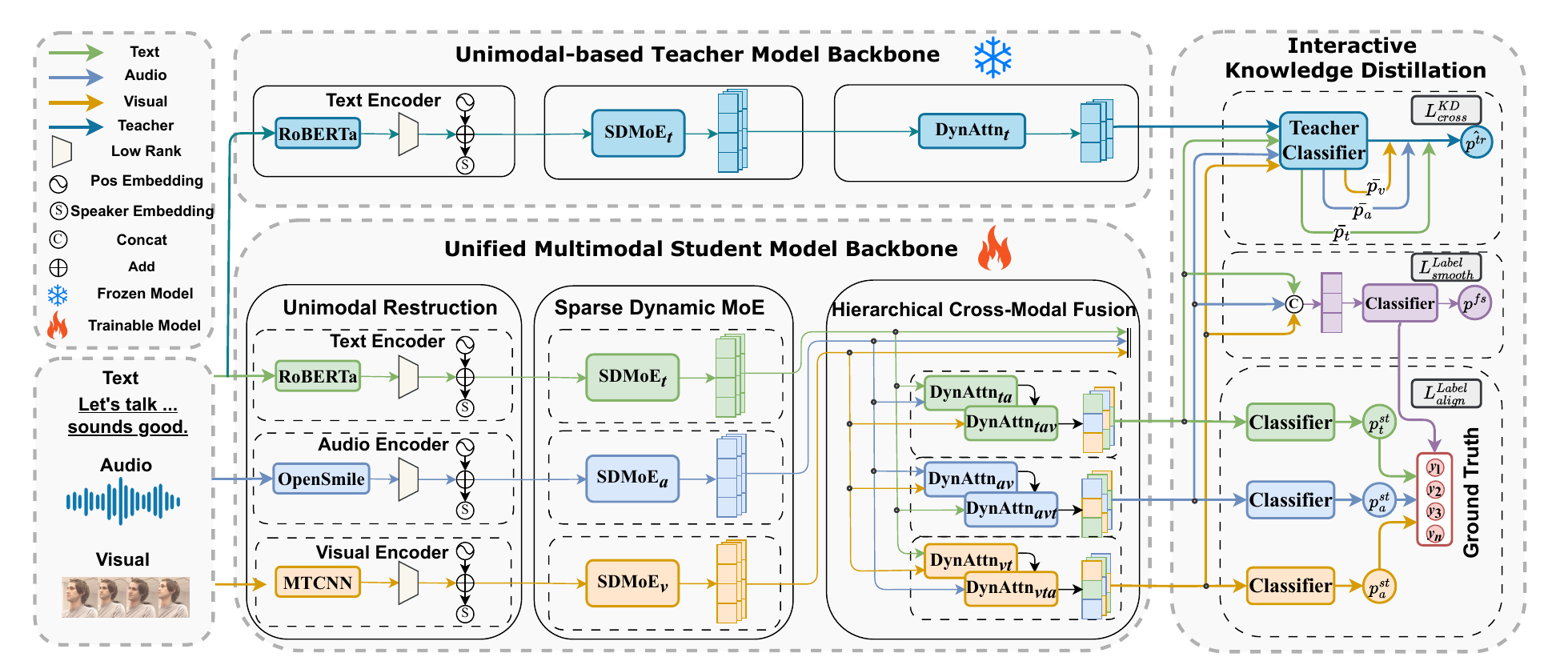}
    \caption{Illustration of the SUMMER framework, where the frozen teacher model is dedicated to mentoring the student model by providing a comprehensive guide for learning with Interactive Knowledge Distillation.}
    \label{fig:SUMMER}
    \end{figure*}

    As illustrated in Fig. \ref{fig:SUMMER}, SUMMER consists of four modules: Unimodal Reconstruction, Sparse Dynamic Mixture of Experts (SDMoE), Hierarchical Cross-Modal Fusion (HCMF), and Interactive Knowledge Distillation (IKD). The unimodal encoder extracts features from three modalities, while SDMoE enhances token-wise interactions by dynamically refining global context associations. HCMF aligns multimodal weights to enrich heterogeneous modal fusion, while IKD facilitates efficient cross-modal fusion through knowledge transfer from a lightweight unimodal-driven pre-trained teacher model.

\subsection{Unimodal Restruction}
    \paragraph{Unimodal Extractor}
    We use RoBERTa model to extract text features $h_i^t \in \mathbb{R}^{l_s \times d_t}$. And then utilize OpenSMILE to extract audio features $h_i^a \in \mathbb{R}^{l_s \times d_a}$. Lastly, we use MTCNN to produce visual features $h_i^v \in \mathbb{R}^{l_s \times d_v}$.

    \paragraph{Utterance-Speaker Embeddings}
    The emotion of the current speaker influences the next speaker's affective state. To model the relationship between speaker identity $s_j$ and utterances, incorporating a latent speaker representation into positional embeddings is essential, where $H^m_i = \{ h^t_i, h^a_i, h^v_i \}$.
    \begin{equation}
        U_e = H_i^m + V_{s_j}o_{s_j} + P_i \in \mathbb{R}^{ls\times d_s},
    \end{equation}
    where $V_{s_j}$ is a learnable speaker embedding, $o_{s_j}$ is the one-hot encoding of each speaker, and $P_i$ represents the absolute position embeddings of the utterance.
    
\subsection{Sparse Dynamic Mixture of Experts}
    In complex conversations, selecting the optimal number of expert models is challenging due to varying contextual intricacies. To mitigate this, we propose SDMoE (Fig. \ref{fig:SDMoE/HCMF} (a)) which adaptively allocates expert models based on intra-modal, ensuring efficient and context-aware processing.
    \paragraph{Auxiliary Expert Network}
        We employ BiGRU experts to capture modality-specific emotional semantics across multiple levels. Each expert encodes corresponding features as $E_{o}=\{BiGRU_{1}(U_{e_{1}}),BiGRU_{2}(U_{e_{2}}),...,BiGRU_{n}(U_{e_{n}})\}$, where $n$ denotes the number of experts.
        
    \paragraph{Dynamic Routing Mechanism}
        To address the limitations of traditional MoE \cite{MoE}, which aggregates weights from fixed Top-K experts, we propose a dynamic routing mechanism $G_{dyn}$, which dynamically adjusts the number of active experts based on task complexity. The gating network generates global expert weights $W_g$, with each scalar $w_g$ indicating the importance of an expert. The recalibrated weights are computed as $G_{dyn} = \frac{Softmax(W_g)}{\tau}$. Weight $w_g$ outside the range $(\mu - \alpha\sigma, \mu + \alpha\sigma)$ are deactivated. Here, temperature $\tau$ controls the smoothness of the distribution. The mean($\mu$) and standard deviation($\sigma$) of $W_g$ define the range for valid weights, with $\alpha$ controlling the selection threshold. A smaller $\alpha$ imposes stricter constraints, reducing noise and redundancy.

        However, our selection mechanism uses discrete sampling, making gradient propagation non-differentiable. To address this, we introduce Gumbel noise $g_{noise} = -log(-log(R_i))$, where $R_i$ is uniformly sampled from $(0,1)$, ensuring differentiability. The refined $\hat{G_{dyn}}$ and the output of the SDMoE module $SD_{sp}$ can be presented as:
        \begin{align}
            \hat{G_{dyn}} &= \frac{\exp(\frac{W_g+g_{noise}}{\tau})}{\sum_1^n\exp(\frac{W_g+g_{noise}}{\tau})}, \\
            SD_{sp} &= \sum_i^n(\hat{G_{dyn}} \times E_o).
        \end{align}

\subsection{Hierarchical Cross-modal Fusion}
    \label{HCMF}
\begin{figure*}[t]
        \centering
        \includegraphics[width=0.85\textwidth]{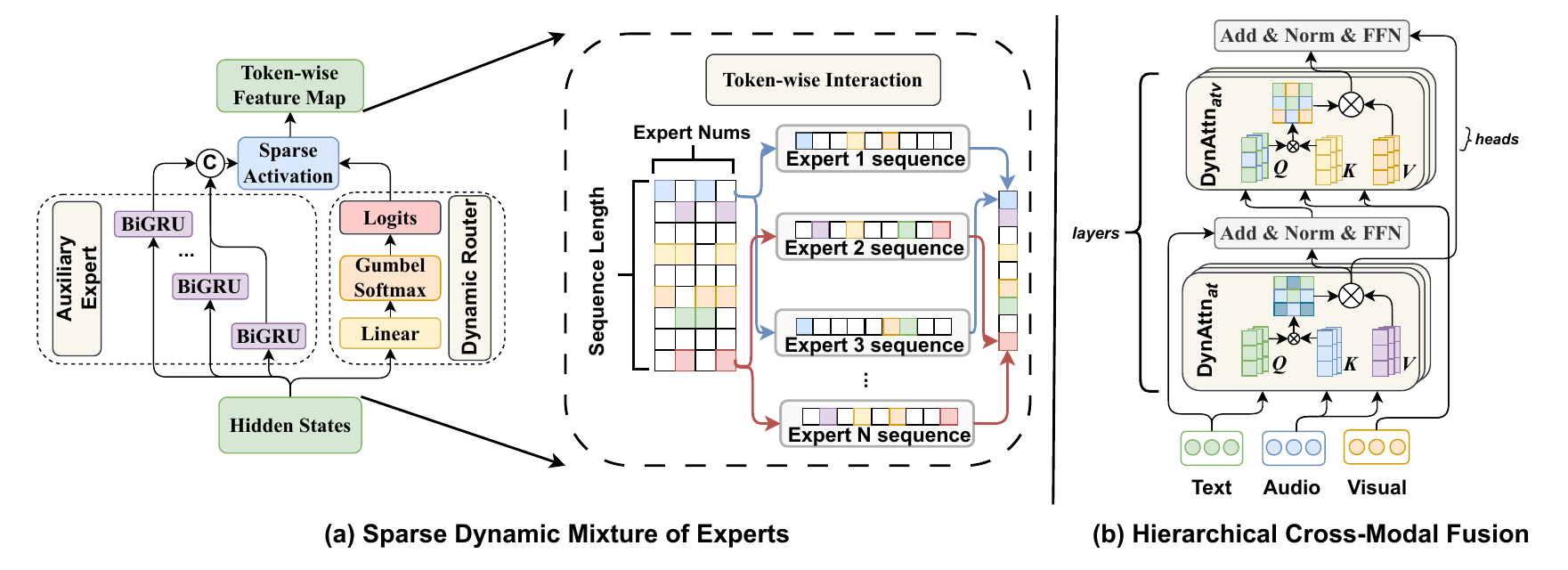}
        \caption{(a) SDMoE comprises two main components: the Auxiliary Expert Network and the Dynamic Routing Mechanism. Specifically, the dynamic router adjusts the relevance of the attention map to facilitate local token-wise interactions. (b) HCMF integrates a multi-level hierarchical structure for cross-modal fusion to enhance overall contextual understanding.}
        \label{fig:SDMoE/HCMF}
    \end{figure*}

    Inter-modal imbalance from heterogeneous hinders stability in multimodal fusion. We propose HCMF framework (Fig. \ref{fig:SDMoE/HCMF} (b)) comprises three branches: HCMF$_t$, HCMF$_a$, and HCMF$_v$, each using a BERT-like encoder for text, audio, and visual.
    
    Taking the HCMF$_t$ branch as an example, the student model inputs are $Q_{st}^t$, $K_{st}^t$, $V_{st}^t$ = $Linear(SD_{sp}^t) \in \mathbb{R}^{l_m \times d_m}$. A learnable factor $\phi$ is applied to adjust the weights of the multi-head attention dynamically. $DynAttn$ is defined as:
        \begin{equation}
            DynAttn = \sum_{i=1}^{n}\phi\cdot Softmax(\frac{Q_{st}^tK_{st}^{t^T}}{\sqrt{d}})V_{st}^t).
         \end{equation}
         
    $DynAttn$ integrates textual, audio, and visual modalities through a three-stage process: (1) learning cross-modal correlations between text (Query) and audio (Key/Value) using bidirectional multi-head cross-attention, (2) fusing text with visual cues using the output from the first stage as Query and visual modality as Key/Value, and (3) applying a feed-forward network for further integration. 
    
    By employing hierarchical $DynAttn$ mechanism, we obtain the fusion outputs $H_{ta}$ (text-audio) and $H_{tav}$ (text-audio-visual), which are formally defined as:
    \begin{equation}
        H_{ta}=DynAttn_{ta}(Q_{st}^t,K_{st}^a,V_{st}^a),
    \end{equation}
    \begin{equation}
        H_{tav}=DynAttn_{tav}(H_{ta},K_{st}^v,V_{st}^v).
    \end{equation}

\subsection{Interactive Knowledge Distillation}
     Gradient disorientation in multimodal learning, driven by heterogeneity, can cause instability. To mitigate this, we pre-trained a unimodal teacher model to guide multimodal learning, using unimodal performance as a prior for cross-modal distillation and transfer learning. 
    \paragraph{Cross KD Loss}
        As shown in prior work on distillation \cite{crosskd}, relying on the teacher's final representations can cause gradient conflicts due to hard labels. Our IKD approach (Fig. \ref{fig:SUMMER}) mitigates this by transferring knowledge via homogeneous probability distributions of heterogeneous modal features. We freeze the teacher modal and apply its classifiers to the student's intermediate features, ensuring a uniform distribution space for heterogeneous modal features. The Cross KD loss $L_{cross}^{KD}$ is computed using KL divergence which defined as:
        \begin{equation}
            L_{cross}^{KD} = \sum_{i=1}^{N} \hat{p^{tr}} \log\frac{\hat{p^{tr}}}{\bar{p^{st}}},
        \end{equation}
        where $N$ denotes the sample size. $\hat{p^{tr}}$ and $\bar{p^{st}}$ represent the predicted distributions of the teacher and student intermediate features, both processed through the teacher model's classifier.
    \paragraph{Align Loss}
        To prevent the student's prediction $p_{i,j}^{st}$ from becoming overly reliant on the teacher model, we constrain the student's training by using the ground truth labels $gt_{i,j}$. Align loss is measured using Cross Entropy loss $L_{align}^{Label}$ and can be represent as follow:
        \begin{equation}
            L_{align}^{Label} =  -\frac{1}{N}\sum_{i=1}^{N}\sum_{j=1}^{C}gt_{i,j}\,log({p_{i,j}^{st}}),
        \end{equation}
        where $C$ denotes the number of emotion categories.
    \paragraph{Label Smooth Loss}
        To reduce noise sensitivity and prevent overconfidence in single categories, we use soft labels instead of hard labels. The smooth loss is defined as:
        \begin{equation}
            L_{smooth}^{Label}=- \frac{1}{N}\sum_{i=1}^N\sum_{j=1}^Cp_{i,j}^{fs}\log(gt_{i,j}),
        \end{equation}
        where $p_{i,j}^{fs}$ represents the prediction through the student model's classifier. For the correct category $gt_{i,j} = 1 - \epsilon$, while for the other categories $gt_{i,j} = \epsilon/ (C - 1)$ where $\epsilon \in (0,1)$.
        
    \paragraph{Training Objectives}
        Our overall training objective of Interactive KD can be represented below, where $\kappa_1,\kappa_2,\kappa_3$ are hyperparameters between different objectives.
        \begin{equation}L_{IKD}=\kappa_1L_{cross}^{KD}+\kappa_2L_{align}^{Label}+\kappa_3 L_{smooth}^{Label}.
        \end{equation}
        
\section{EXPERIMENTAL SETTINGS}
\subsection{Datasets and Evaluation Metrics}
    We evaluate SUMMER on two benchmark MERC datasets, IEMOCAP \cite{iemocap} and MELD \cite{meld}, which include multimodal data (text, audio, video). IEMOCAP contains 12 hours of annotated conversations with six emotion labels, while MELD features dialogue clips from Friends with seven labels. Performance is measured using accuracy (Acc), F1-score, weighted accuracy (w-Acc), and weighted F1-score (w-F1) to compare SUMMER with baseline methods.

\subsection{Baselines}
    We compare our model with strong baselines: DialogueRNN \cite{dialoguernn} and DialogueGCN \cite{dialoguegcn} use GRUs and GCNs for conversational modeling, respectively. MMGCN \cite{mmgcn} and CORECT \cite{CORECT} integrate GCNs with dynamic fusion for multimodal context, while MultiEMO \cite{MultiEMO} employs correlation-aware attention. SDT \cite{SDT} leverages self-distillation for intra- and inter-modal interactions, and CHFusion \cite{CHFusion} restructures contextual information via hierarchical fusion. Additionally, we evaluate against the latest LLM-based model, InstructERC \cite{InstructERC}.

\subsection{Implementation Details}
    The model is implemented in PyTorch, employing the Adam with learning rates $lr=1e-4$, and batch sizes $bs=32$ and $bs=100$ for IEMOCAP and MELD. To prevent overfitting, modal input dimensions are configured as 100 (text/audio) and 256 (visual) for IEMOCAP, and 768 (text), 512 (audio), and 1000 (visual) for MELD. The HCMF architecture features a hidden size of 1024, 4 attention heads, and 6 cross-modal fusion layers, with an L2 weight decay of 1e-5. Loss weight hyperparameters are configured as $\kappa_1=0.4, \kappa_2=0.3, \kappa_3=0.3$, while the temperature $\tau=0.5$, controlling factor $\alpha=2$, and the smoothing parameter $\epsilon=0.1$. 
    \begin{table}[]
            \centering
            \captionof{table}{Ablation study with different modality settings.}
            \fontsize{7}{7}\selectfont
            \renewcommand{\arraystretch}{1.5}
            % \vspace{-0.12in}
            \setlength{\tabcolsep}{1.0mm}
            \begin{tabular}{l|cc|cc}
            \hline
            \multicolumn{1}{c|}{\multirow{2}{*}{Modality}} & \multicolumn{2}{c|}{IEMOCAP}& \multicolumn{2}{c}{MELD}\\ \cline{2-5}
            \multicolumn{1}{c|}{}& ACC& w-F1& ACC& w-F1\\ \hline
            Text& 69.57& 69.73& 66.49& 65.32\\
            Audio& 67.37& 67.18& 55.78& 55.47\\
            Visual& 66.20& 66.28& 53.89& 53.43\\
            Text+Audio& 71.18& 70.83& 67.55& 66.58\\
            Text+Visual& 69.80& 69.51& 67.54& 66.41\\
            Audio+Visual& 68.05& 67.49& 59.01& 58.33\\
            Text+Audio+Visual& \textbf{71.62} & \textbf{71.18} & \textbf{67.71} & \textbf{66.61} \\ \hline
            \end{tabular}
            \label{tab:modality_combination_results}
    \end{table}
    \begin{table*}[]
        \centering
        \caption{Quantative comparisons on IEMOCAP(6-ways) multimodal (A+V+T) setting.}
        % \captionsetup{belowskip=1pt}
        \fontsize{7}{7}\selectfont
        \setlength{\tabcolsep}{1.0mm}
        \renewcommand{\arraystretch}{1.5}
        \begin{tabular}{l|cc|cc|cc|cc|cc|cc|cc}
        \hline
        \multicolumn{1}{c|}{\multirow{2}{*}{Models}} & \multicolumn{2}{c|}{happy}& \multicolumn{2}{c|}{sad}& \multicolumn{2}{c|}{neutral}&\multicolumn{2}{c|}{anger}& \multicolumn{2}{c|}{excitement} & \multicolumn{2}{c|}{frustration} & \multirow{2}{*}{w-ACC} & \multirow{2}{*}{w-F1} \\ \cline{2-13}
        \multicolumn{1}{c|}{}& ACC& F1& ACC& F1& ACC& F1& ACC& F1& ACC& F1& ACC& F1&&\\ \hline
        DialogueRNN& 44.05& 32.46& \textbf{86.61} & 82.73& 54.08& 54.64& 67.72& 65.24& 63.71& 70.64& 56.23& 57.11& 61.81& 61.55\\
        DialogueGCN& 61.11& 51.87& 84.90& 76.76& 69.27& 56.76& \textbf{76.47} & 62.26& 76.25& 72.71& 50.39& 58.04& 69.73& 63.07\\
        MMGCN& 48.94& 38.66& 80.54& 76.39& 59.56& 61.73& 74.68& 68.18& 71.91& 74.80& 60.53& 62.97& 65.87& 65.67\\
        CORECT& 59.15& 58.74& {\ul 86.18}& 80.95& 71.43& 69.52& 63.74& 65.91& 80.60& 76.19& 62.89& 68.11& 71.44& 70.81\\
        MultiEMO& 53.80& 56.29& 83.95& 80.18& 75.84& 69.76& 67.86& 67.46& 79.78& 76.01& 64.40& 69.42& 72.31& 71.64\\
        SDT& 61.96& 65.80& 85.46& 82.20& 76.16& 72.70& 63.27& 67.76& 78.12& {\ul 82.94}    & 64.51& 67.90& 74.44& 74.13\\
        CHFusion& -& -& -& -& -& -& -& -& -& -& -& -& {\ul 76.50}& {\ul 76.80}\\
        InstructERC& -& -& -& -& -& -& -& -& -& -& -& -& -& {\ul 71.39}\\ \hline
        Teacher Model& {\ul 70.83}& {\ul 73.12}& 82.79& {\ul 83.61}& \textbf{84.86} & {\ul 74.23}& 65.22& {\ul 71.95}& {\ul 82.94}& 81.30& {\ul 68.63}& {\ul 70.10}    & 75.21& 74.22\\
        Student Model& \textbf{71.72} & \textbf{74.29} & 82.52& \textbf{85.47} & {\ul 78.45}& \textbf{80.46} & {\ul 75.97}& \textbf{72.67} & \textbf{88.76} &\textbf{84.34} & \textbf{73.94}  & \textbf{73.42} & \textbf{79.11}&\textbf{78.95}\\ \hline
        \end{tabular}
        \label{IEMOCAP_results}
    \end{table*}
    \begin{table*}[]
    \centering
        \caption{Quantative comparisons on MELD(7-ways) multimodal (A+V+T) setting.}
        \fontsize{7}{7}\selectfont
        \setlength{\tabcolsep}{1.0mm}
        \renewcommand{\arraystretch}{1.5}
        \begin{tabular}{l|cc|cc|cc|cc|cc|cc|cc|cc}
        \hline
        \multicolumn{1}{c|}{\multirow{2}{*}{Models}} & \multicolumn{2}{c|}{neutral}& \multicolumn{2}{c|}{surprise}& \multicolumn{2}{c|}{fear}& \multicolumn{2}{c|}{sadness}& \multicolumn{2}{c|}{joy}& \multicolumn{2}{c|}{disgust}&\multicolumn{2}{c|}{anger}& \multirow{2}{*}{w-ACC$\uparrow$} & \multirow{2}{*}{w-F1$\uparrow$} \\ \cline{2-15}
        \multicolumn{1}{c|}{}& ACC& F1& ACC& F1& ACC& F1& ACC& F1& ACC& F1& ACC& F1& ACC& F1&&\\ \hline
        MMGCN& 68.87& 77.51& 48.12& 46.80& 0& 0& 50.00& 13.33& 55.46& 51.47& 0& 0& 45.40& 45.60& 56.85& 57.35\\
        DialogueRNN& 71.62& 75.66& 52.17& 46.97& 0& 0& 32.46& 22.98& 48.00& 52.00& 0& 0& 43.60& 45.88& 55.83& 57.37\\
        DialogueGCN& 79.06& 75.80& 53.02& 50.42& 0& 0& 17.79& 23.72& 59.20& 55.48& 0& 0& 50.43& 48.27& 60.96& 58.72\\
        CORECT& 80.00& {\ul 81.60}& 58.49& 49.60& 37.90& 26.47& 52.53& {\ul 43.78}&\textbf{67.79} & 63.32& 44.83& 31.58& 52.72& 51.64& 66.01& 65.92\\
        SDT& 76.96& 79.85& 56.75& 57.54& 25.00& 17.95& \textbf{58.20} & 43.03& 65.72& 64.56& 39.47& 28.30& 50.64& 53.80& 66.10& 66.19\\
        MultiEMO& 78.55& 79.94& 54.49& 58.28& 36.00& 24.00& {\ul 56.15}& 43.20& 61.06& 64.64& 43.75& 28.00& {\ul 53.31}& 53.47& 66.43& 66.40\\
        InstructERC& -& -& -& -& -& -& -& -& -& -& -& -& -& -& -& 69.15\\ \hline
        Teacher Model& {\ul 82.78}& 76.92& \textbf{62.70} & {\ul 65.35}& {\ul 52.80}& {\ul 55.74}& 49.37& \textbf{45.66} & 65.13& {\ul 69.03}& \textbf{45.37} & {\ul 45.04}& 52.44& {\ul 56.59}& {\ul 66.92}& {\ul 67.59}\\
        Student Model& \textbf{86.29} & \textbf{83.44} & {\ul 62.66}& \textbf{68.95} & \textbf{53.42} & \textbf{56.39} & 49.38& 43.04& {\ul 66.86}& \textbf{70.96} & {\ul 45.28}& \textbf{47.52} & \textbf{55.13} & \textbf{57.33} & \textbf{68.78}& \textbf{69.81}\\ \hline
        \end{tabular}
        \label{MELD_results}
    \end{table*}
     
\subsection{Results and Analysis}    
    Tables \ref{IEMOCAP_results} and \ref{MELD_results} compare performance metrics on IEMOCAP and MELD with baseline models, which results validate SUMMER's ability to handle class imbalance and achieve fine-grained emotional distinctions.
    
     On IEMOCAP, SUMMER improves w-ACC by 2.61\% and w-F1 by 2.15\%, outperforming baselines like CHFusion. The teacher model achieves notable gains over most of the methods, while the student model performs best in categories such as ``happy'' (w-ACC +9.76\%, w-F1 +8.49\%), ``excitement'' (w-ACC +10.64\%), and ``frustration'' (w-ACC +9.43\%, w-F1 +5.52\%) compared to SDT. On MELD, the teacher model performs stable, while the student model excels in underrepresented emotions, improving ``fear'' (w-ACC +17.42\%), ``anger'' (w-F1 +3.86\%) and ``disgust'' (w-F1 +19.52\%)) over MultiEMO. Furthermore, SUMMER exceeds the LLM-based model InstructERC, highlighting the benefits of dynamic fusion and retrograde distillation strategy in intra- and inter-modal.
    
\subsection{Ablation Studies}
\label{sec:Ablation_Studies}
    To evaluate the contribution of each SUMMER component, we perform ablation studies on the IEMOCAP and MELD datasets, with results shown in Tables \ref{tab:modality_combination_results} and \ref{SUMMER_ablation}.
    \paragraph{Guidelines for Teacher Model Selection}
        We test various modality combinations (text, audio, visual) using the original attention mechanism. Table \ref{tab:modality_combination_results} shows that the text modality consistently outperforms others, justifying its selection as the teacher model. Although combining text with other modalities yields minor gains, the increased model complexity and overfitting risks favor a unimodal teacher for efficiency.
        \begin{table}[t]
            \centering
            \captionof{table}{Ablation study of key components on IEMOCAP and MELD.}
            \fontsize{7}{7}\selectfont
            \renewcommand{\arraystretch}{1.5}
            % \vspace{-0.12in}
            \setlength{\tabcolsep}{1.0mm}
            \begin{tabular}{lll|cc|cc}
            \hline
            \multicolumn{3}{c|}{Module}& \multicolumn{                2}{c|}{IEMOCAP} & \multicolumn{2}{c}{MELD} \\ \hline
            \multicolumn{1}{c}{SDMoE} & \multicolumn{1}{c}{HCMF} & \multicolumn{1}{c|}{IKD} & ACC& w-F1& ACC& w-F1\\ \hline
            $\times$& $\times$& $\times$& 74.44& 74.13& 66.10& 66.19\\
            $\checkmark$& $\times$& $\times$& 76.52(+2.08)& 76.64(+2.51)& 67.43(+1.33)& 68.57(+2.38)\\
            $\times$& $\checkmark$& $\times$& 76.15(+1.71)& 75.43(+1.30)& 67.83(+1.73)& 68.24(+2.05)\\
            $\times$& $\times$& $\checkmark$& 77.48(+3.04)& 76.86(+2.73)& 68.39(+2.29)& 69.52(+3.33)\\ \hline
            $\checkmark$& $\checkmark$& $\times$& 77.82(+3.38)& 78.13(+3.99)& 68.17(+2.07)& 69.04(+2.85)\\
            $\times$& $\checkmark$& $\checkmark$& 77.95(+3.51)& 77.94(+3.81)& 68.42(+2.32)& 69.21(+3.02)\\
            $\checkmark$& $\times$& $\checkmark$& 78.54(+4.10)& 78.64(+4.51)& 68.57(+2.47)& 69.33(+3.14)\\ \hline
            $\checkmark$& $\checkmark$& $\checkmark$& \textbf{79.11(+4.69)}& \textbf{78.95(+4.82)}& \textbf{68.78(+2.68)}& \textbf{69.81(+3.62)}\\ \hline
            \end{tabular} 
            \label{SUMMER_ablation}
        \end{table}
    \paragraph{Validity of SDMoE modules}
        Adding SDMoE (Table \ref{SUMMER_ablation}) reveals consistent performance increase (Row 2) across emotion categories and achieves significant gains over benchmarks. SDMoE dynamically adjusts the token-wise selection and allocation, which minimizes redundancy and enhances overall intra-modal performance.
    % \begin{figure}[t]
    %     \centering
    %     \includegraphics[width=\linewidth]{figs_official/w_f1_scores_comparison_two_charts_horizontal.pdf}
    %     \caption{Performance of the SDMoE module across various modalities on the IEMOCAP and MELD datasets.}
    %     \label{fig:barchart}
    % \end{figure}
    \paragraph{Impact of HCMF}
        Ablation experiments by appending the HCMF module revealed a notable performance improvement (Row 3), confirming that HCMF surpasses static fusion strategies in integrating multimodal information and enhancing the model's ability to learn high-level semantic representations within inter-modal.
    \paragraph{Interactive Knowledge Distillation}
        As shown in Table \ref{SUMMER_ablation}, the novel IKD achieves optimal performance (Row 4) by guiding the student model with frozen teacher representations, enhancing its ability to integrate inter-modal relationships. Soft labels improve generalization by preserving relational information, while KL divergence stabilizes training and mitigates gradient conflicts from modality heterogeneity.
        
    By combining three different modules, we observe a corresponding improvement (Row 5, 6, 7) in model performance, with the optimal results (Row 8) achieved through the simultaneous integration of all three modules.
    \paragraph{Error Analysis}
        In our study, the teacher model effectively captures fine-grained unimodal features, while the student model leverages multimodal fusion for more generalized and robust performance. Despite slightly lower performance in certain categories, such as ``sad'', the student model performs well overall. The underperformance may stem from multimodal conflicts, overlapping emotional boundaries (e.g., sadness and frustration), and data imbalance, which need addressing to enhance multimodal emotion recognition.

\subsection{Multi-modal Representation Visualization}
\begin{figure}[t]
        \centering
        \includegraphics[width=0.9\linewidth]{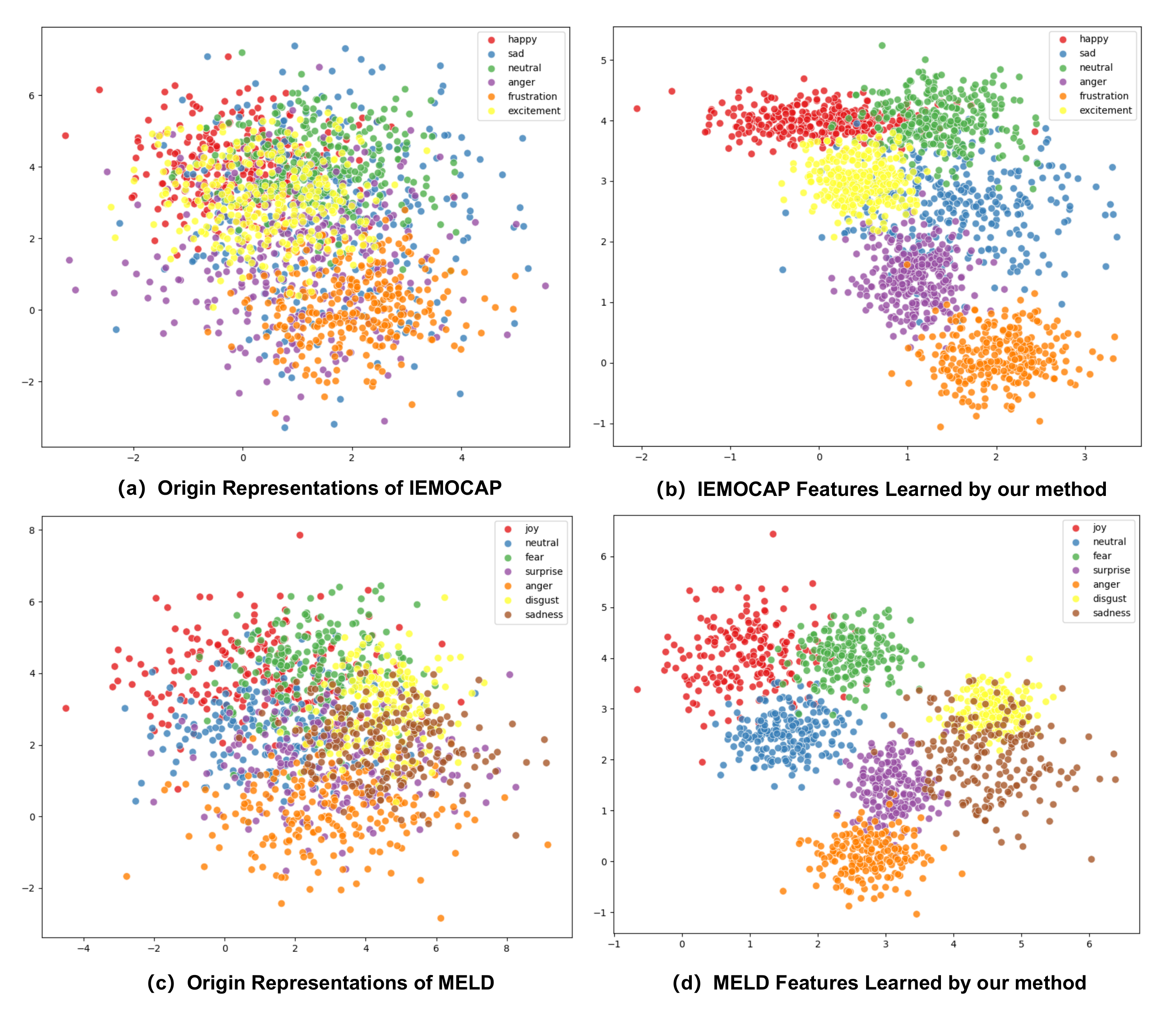}
        \caption{Visualization of features where each point corresponds to an utterance, with colors denoting different emotions.}
        \label{fig:scatter}
    \end{figure}

    To evaluate our method, we applied t-SNE to project high-dimensional multimodal features into a two-dimensional space (Fig. \ref{fig:scatter}). The proposed SUMMER model achieves a clear separation of emotion categories with minimal overlap, particularly for similar emotions like ``happy'' and ``excitement'', while enhancing clustering and distinctions between neutral and other emotions. Additionally, SUMMER demonstrates robustness in multimodal integration, effectively capturing subtle emotional variations despite noise and blurred boundaries.
\section{Conclusion}
    In this work, we propose SUMMER framework for MERC tasks, effectively integrating heterogeneous modalities through a Sparse Dynamic Mixture of Experts for local token-wise interaction combined with the Hierarchical Cross-modal Fusion. By employing a novel interactive Knowledge Distillation where a unimodal teacher guides a multimodal student model, SUMMER mitigates gradient disorientation learning and enhances intra- and inter-modal learning. Experiments on IEMOCAP and MELD datasets show that SUMMER outperforms state-of-the-art methods, improving recognition of both majority and minority emotion classes, and highlighting its robustness in MERC tasks.

\bibliographystyle{IEEEbib}
\bibliography{main}

\begin{thebibliography}{10}

\bibitem{MultiEMO}
Tao Shi and Shao-Lun Huang,
\newblock ``Multiemo: An attention-based correlation-aware multimodal fusion
  framework for emotion recognition in conversations,''
\newblock in {\em Proceedings of the 61st Annual Meeting of the Association for
  Computational Linguistics (Volume 1: Long Papers)}, 2023, pp. 14752--14766.

\bibitem{SDT}
Hui Ma, Jian Wang, Hongfei Lin, Bo~Zhang, Yijia Zhang, and Bo~Xu,
\newblock ``A transformer-based model with self-distillation for multimodal
  emotion recognition in conversations,''
\newblock {\em IEEE Transactions on Multimedia}, 2023.

\bibitem{CTNet}
Zheng Lian, Bin Liu, and Jianhua Tao,
\newblock ``Ctnet: Conversational transformer network for emotion
  recognition,''
\newblock {\em IEEE/ACM Transactions on Audio, Speech, and Language
  Processing}, vol. 29, pp. 985--1000, 2021.

\bibitem{CKETF}
Soumitra Ghosh, Deeksha Varshney, Asif Ekbal, and Pushpak Bhattacharyya,
\newblock ``Context and knowledge enriched transformer framework for emotion
  recognition in conversations,''
\newblock in {\em 2021 International Joint Conference on Neural Networks
  (IJCNN)}. IEEE, 2021, pp. 1--8.

\bibitem{emocaps}
Zaijing Li, Fengxiao Tang, Ming Zhao, and Yusen Zhu,
\newblock ``Emocaps: Emotion capsule based model for conversational emotion
  recognition,''
\newblock {\em ArXiv Preprint ArXiv:2203.13504}, 2022.

\bibitem{tailor}
Yi~Zhang, Mingyuan Chen, Jundong Shen, and Chongjun Wang,
\newblock ``Tailor versatile multi-modal learning for multi-label emotion
  recognition,''
\newblock in {\em Proceedings of the AAAI Conference on Artificial
  Intelligence}, 2022, vol.~36, pp. 9100--9108.

\bibitem{SENet}
Samuel Albanie, Arsha Nagrani, Andrea Vedaldi, and Andrew Zisserman,
\newblock ``Emotion recognition in speech using cross-modal transfer in the
  wild,''
\newblock in {\em Proceedings of the 26th ACM International Conference on
  Multimedia}, 2018, pp. 292--301.

\bibitem{Schoneveld}
Liam Schoneveld, Alice Othmani, and Hazem Abdelkawy,
\newblock ``Leveraging recent advances in deep learning for audio-visual
  emotion recognition,''
\newblock {\em Pattern Recognition Letters}, vol. 146, pp. 1--7, 2021.

\bibitem{fasd}
Rui Yu, Runkai Zhao, Jiagen Li, Qingsong Zhao, Songhao Zhu, HuaiCheng Yan, and
  Meng Wang,
\newblock ``Unleashing the potential of mamba: Boosting a lidar 3d sparse
  detector by using cross-model knowledge distillation,''
\newblock {\em arXiv preprint arXiv:2409.11018}, 2024.

\bibitem{MoE}
Albert~Q Jiang, Alexandre Sablayrolles, Antoine Roux, Arthur Mensch, Blanche
  Savary, Chris Bamford, Devendra~Singh Chaplot, Diego de~las Casas, Emma~Bou
  Hanna, Florian Bressand, et~al.,
\newblock ``Mixtral of experts,''
\newblock {\em arXiv preprint arXiv:2401.04088}, 2024.

\bibitem{crosskd}
Jiabao Wang, Yuming Chen, Zhaohui Zheng, Xiang Li, Ming-Ming Cheng, and Qibin
  Hou,
\newblock ``Crosskd: Cross-head knowledge distillation for object detection,''
\newblock in {\em Proceedings of the IEEE/CVF Conference on Computer Vision and
  Pattern Recognition}, 2024, pp. 16520--16530.

\bibitem{iemocap}
Carlos Busso, Murtaza Bulut, Chi-Chun Lee, Abe Kazemzadeh, Emily Mower, Samuel
  Kim, Shrikanth~S Narayanan, et~al.,
\newblock ``Iemocap: Interactive emotional dyadic motion capture database,''
\newblock {\em Language Resources and Evaluation}, vol. 42, pp. 335--359, 2008.

\bibitem{meld}
Soujanya Poria, Devamanyu Hazarika, Navonil Majumder, Gautam Naik, Erik
  Cambria, and Rada Mihalcea,
\newblock ``Meld: A multimodal multi-party dataset for emotion recognition in
  conversations,''
\newblock {\em ArXiv Preprint ArXiv:1810.02508}, 2018.

\bibitem{dialoguernn}
Navonil Majumder, Soujanya Poria, Devamanyu Hazarika, Rada Mihalcea, Alexander
  Gelbukh, and Erik Cambria,
\newblock ``Dialoguernn: An attentive rnn for emotion detection in
  conversations,''
\newblock in {\em Proceedings of the AAAI Conference on Artificial
  Intelligence}, 2019, vol.~33, pp. 6818--6825.

\bibitem{dialoguegcn}
Deepanway Ghosal, Navonil Majumder, Soujanya Poria, Niyati Chhaya, and
  Alexander Gelbukh,
\newblock ``Dialoguegcn: A graph convolutional neural network for emotion
  recognition in conversation,''
\newblock {\em ArXiv Preprint ArXiv:1908.11540}, 2019.

\bibitem{mmgcn}
Jingwen Hu, Yuchen Liu, Jinming Zhao, and Qin Jin,
\newblock ``Mmgcn: Multimodal fusion via deep graph convolution network for
  emotion recognition in conversation,''
\newblock {\em ArXiv Preprint ArXiv:2107.06779}, 2021.

\bibitem{CORECT}
Cam-Van~Thi Nguyen, Anh-Tuan Mai, The-Son Le, Hai-Dang Kieu, and Duc-Trong Le,
\newblock ``Conversation understanding using relational temporal graph neural
  networks with auxiliary cross-modality interaction,''
\newblock {\em ArXiv Preprint arXiv:2311.04507}, 2023.

\bibitem{CHFusion}
Navonil Majumder, Devamanyu Hazarika, Alexander Gelbukh, Erik Cambria, and
  Soujanya Poria,
\newblock ``Multimodal sentiment analysis using hierarchical fusion with
  context modeling,''
\newblock {\em Knowledge-based Systems}, vol. 161, pp. 124--133, 2018.

\bibitem{InstructERC}
Shanglin Lei, Guanting Dong, Xiaoping Wang, et~al.,
\newblock ``Instructerc: Reforming emotion recognition in conversation with a
  retrieval multi-task llms framework,''
\newblock {\em arXiv preprint arXiv:2309.11911}, 2023.

\end{thebibliography}

\end{document}